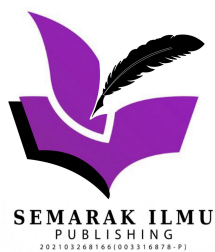

# Semarak International Journal of Applied Psychology

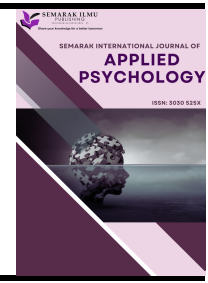

# Enhanced Suicidal Ideation Detection from Social Media using a CNN-BiLSTM Hybrid Model

Mohaiminul Islam Bhuiyan[1], Nur Shazwani Kamarudin[1,*], Nur Hafieza Ismail[1], Jarin Tasnim Raya[2]

[1] Faculty of Computing, Universiti Malaysia Pahang Al-Sltan Abdullah, 26600, Pekan, Pahang, Malaysia
[2] Department of Computer Science and Engineering, University of Asia Pacific (UAP), Dhaka 1205, Bangladesh

| ARTICLE INFO | ABSTRACT |
|---|---|
|  | Suicidal ideation detection is crucial for preventing suicides, a leading cause of death worldwide. Many individuals express suicidal thoughts on social media, offering a vital opportunity for early detection through advanced machine learning techniques. The identification of suicidal ideation in social media text is improved by utilising a hybrid framework that integrates Convolutional Neural Networks (CNN) and Bidirectional Long Short-Term Memory (BiLSTM), enhanced with an attention mechanism. To enhance the interpretability of the model's predictions, Explainable AI (XAI) methods are applied, with a particular focus on SHapley Additive exPlanations (SHAP). At first, the model managed to reach an accuracy of 92.81%. By applying fine-tuning and early stopping techniques, the accuracy improved to 94.29%. The SHAP analysis revealed key features influencing the model's predictions, such as terms related to mental health struggles. This level of transparency boosts the model's credibility while helping mental health professionals understand and trust the predictions. This work highlights the potential for improving the accuracy and interpretability of detecting suicidal tendencies, providing a foundation for future research in transparent mental health AI systems. It emphasizes the significance of blending powerful machine learning methods with explainability to develop reliable and impactful mental health solutions. |

## 1. Introduction

With the ongoing transformation of society through digital technologies, social media has become a key space for observing users' psychological states. How people communicate and express themselves on these platforms provides valuable insights into their mental health. One of the most pressing concerns revealed through online expressions is suicidal ideation, a growing public health issue that demands immediate attention. The World Health Organization (WHO) reports that around 800,000 individuals lose their lives to suicide each year [1], making it one of the primary causes of death worldwide. It is especially concerning that suicide ranks as the second most common cause of death among people aged 15 to 29. Suicide arises from a variety of complex factors, frequently associated with mental health conditions like depression, substance abuse, and psychosis, though these factors are complex and differ from person to person. Many people suffering from suicidal thoughts faces a lack of support, and traditional methods of identifying suicidal ideation-such as self-reporting-are often unreliable due to fear of stigma or hospitalization, leading individuals to hide their true feelings.

---

* *Corresponding author.*
*E-mail address: t nshazwani@umpsa.edu.my*

A promising approach for detecting suicidal ideation has emerged using digital platforms, particularly social media, where individuals may express their struggles. The vast amount of data produced on these platforms serves as an asset for predictive modeling. Advances in Machine Learning (ML) and Deep Learning (DL) [2] have made it possible to analyze social media posts for signs of suicidal ideation. Technologies such as Convolutional Neural Networks (CNN) [3], Recurrent Neural Networks (RNN), and hybrid models can detect subtle linguistic patterns and complex meanings in text that were once challenging to identify [4]. Despite their accuracy, many of these models function as "black boxes," meaning they make predictions without offering clear explanations for how decisions were reached [5]. The absence of transparency poses a significant challenge in clinical settings, where knowing why a model makes a certain prediction is just as crucial as the prediction itself. To overcome this, Explainable AI (XAI) techniques have become vital, offering clearer insights into the factors that shape a model's decisions and making AI systems more understandable and trustworthy [6]. This added layer of transparency is particularly critical for healthcare and mental health applications, where trust in the model's output is vital for clinical use.

This study presents several significant contributions aimed at enhancing the detection of suicidal ideation through social media texts. Firstly, a hybrid model is presented that merges Convolutional Neural Networks (CNN) with Bidirectional Long Short-Term Memory (BiLSTM) layers, incorporating an attention mechanism for enhanced performance. This hybrid architecture efficiently captures both fine-grained local details and broader global patterns in the text, enabling a more detailed analysis of social media posts. Furthermore, model optimization is achieved through the implementation of fine-tuning and early stopping techniques, which resulted in an improvement in performance. These optimization strategies proved crucial in enhancing accuracy while mitigating the risk of overfitting.

Another key contribution is the use of Explainable AI (XAI) techniques, which provide transparency and interpretability to the model's predictions. This is especially crucial in clinical environments, where gaining insight into the model's decision-making process is vital for fostering trust and ensuring dependable outcomes. Lastly, a thorough comparative analysis is conducted, showing that the proposed model surpasses current machine learning (ML) and deep learning (DL) methods in accuracy, precision, recall, and F1 score. The results underscore the model's greater effectiveness in identifying signs of suicidal ideation from social media content.

In recent years, there has been a growing focus on identifying suicidal thoughts through social media, driven by the increasing rates of suicide worldwide. Researchers have employed various machine learning (ML) [7] and deep learning (DL) techniques to tackle this issue, achieving promising results. Aldhyani et al. trained on 232,074 Reddit SuicideWatch posts and compared a CNN–BiLSTM with XGBoost. Using raw textual features, CNN–BiLSTM reached 95% accuracy (XGBoost 91.5%); with LIWC features, XGBoost performed better (86.9% vs 84.5%), showing trees are competitive on lexicon features while deep hybrids excel on raw text [8]. Jain et al. used the Random Forest (RF) method to distinguish between suicidal and non-suicidal texts. By analyzing a dataset of social media posts, their model reached an accuracy of 77.298%. Though the accuracy was lower than that of other models, the study emphasized the crucial role of feature selection and preprocessing in enhancing overall model performance [9]. Tadesse et al. created a model that integrated Convolutional Neural

Networks (CNN) with Long Short-Term Memory (LSTM) networks. The model was trained on data from Reddit, specifically targeting subreddits discussing mental health topics. With an accuracy of 93.8%, the results highlighted the strength of hybrid models in recognizing both short-term patterns and long-term dependencies within text data [10].

Sawhney et al. explored the use of transformer-based models combined with Bidirectional Long Short-Term Memory (BiLSTM) networks. Using a comprehensive dataset from multiple social media platforms, they achieved an accuracy of 87.34%. This study showcased the potential of transformer architectures in enhancing the contextual understanding of suicidal ideation expressions [11]. Naghavi et al. employed an ensemble of decision trees (DT) to detect suicidal ideation from social media posts. Their model, trained on a dataset comprising various mental health-related subreddits, achieved an accuracy of 93%. The study emphasized the robustness of ensemble methods in handling diverse and noisy text data [12]. De Choudhury et al. analyzed Twitter data to identify users with suicidal ideation using linguistic features and social engagement metrics. Their approach provided valuable insights into the behavioral patterns of individuals expressing suicidal thoughts on social media [13]. Burnap et al. applied sentiment analysis and machine learning techniques to classify tweets related to suicidal behavior. They used a large dataset from Twitter and achieved significant results in early detection, underscoring the importance of real-time data analysis in preventing suicides [14].

Coppersmith *et al.*, [15] carried out a study that used linguistic analysis along with machine learning methods to identify signs of suicidal ideation in social media posts. They used data from Twitter and demonstrated that specific linguistic markers could effectively indicate suicidal tendencies, achieving an accuracy of 85% [15]. Sawhney et al. evaluated a C-LSTM (CNN→LSTM) model for suicidal-ideation detection on Twitter. They first built a suicide-lexicon from web forums (with Tumblr/Reddit posts aiding term discovery), then collected and manually annotated 5,213 tweets; 822 (15.76%) were labeled suicidal by three annotators. Their C-LSTM outperformed RNN/LSTM and linear baselines, reaching 0.812 accuracy and 0.827 F1 with 10-fold cross-validation—evidence that combining local n-gram features from CNN with the LSTM's sequence modeling improves prediction [16]. Li *et al.*, [17] introduced BLAM, a BERT + BiLSTM model trained with multi-task learning (auxiliary emotion classification) and adversarial training for robustness. Evaluated on a COVID-era Twitter corpus for suicidal-ideation detection, BLAM outperformed strong baselines, highlighting how transformer context plus sequence modeling and the added signal from emotions boosts detection performance on noisy social media text [17]. Matero et al. designed a model that combined hierarchical attention networks with LSTM to identify suicidal ideation. Their study utilized data from Reddit and Twitter, achieving an accuracy of 88.7%. The hierarchical attention mechanism allowed their model to focus on different levels of textual information, improving interpretability and performance [18]. Buddhitha and Inkpen present an MTL model that jointly learns suicidality and mental-disorder detection with shared and task-specific layers. Trained on UMD (Reddit) with SMHD and validated on CLPsych-2015 (Twitter), their CNN-based MTL achieves state-of-the-art macro-F1 on both "flagged" and "urgent" tasks. Signals from comorbid disorders - especially PTSD - consistently boost suicidality detection and improve cross-platform generalization [19].

These studies collectively demonstrate the progress in using machine learning and deep learning methods to identify suicidal thoughts. However, existing approaches for suicidal ideation detection operate as 'black boxes' without providing interpretable explanations for their predictions, which hinders clinical adoption and trust. Additionally, many models struggle to achieve the high accuracy levels necessary for reliable real-world deployment in mental health monitoring. The use of hybrid models, along with methods that improve interpretability, as shown in the proposed CNN-BiLSTM model, boosts both accuracy and clarity. This makes the systems more dependable for real-world use.

## 2. Methodology
*2.1 Dataset*

The dataset for this study was obtained from the Kaggle data repository and includes 232,074 samples in total. This dataset includes texts labeled as suicidal and non-suicidal, with an equal distribution of 116,037 samples in each category. The data for this study was gathered from Reddit's "SuicideWatch" and "depression" subreddits, covering posts made between December 16, 2008, and January 2, 2021, for "SuicideWatch," and from January 1, 2009, to January 2, 2021, for "depression." To ensure a balanced dataset for training and evaluation, non-suicidal posts were sourced from the r/teenagers subreddit.

*2.2 Data Pre-Processing*

Preparing data is a vital part of developing models, especially when handling text from social media. This type of data frequently includes noise, informal language, and irregularities that can affect the performance and accuracy of the model [20]. Proper preprocessing helps to clean and structure the data, ensuring better results and more reliable outcomes. The following preprocessing steps were undertaken:

Tokenization: Tokenization involves breaking down large sections of text into smaller pieces known as tokens. Depending on the method used, these tokens can represent individual words, parts of words, or even single characters. This process helps structure the text for easier analysis and interpretation. In the context of this experiment, tokenization was performed on the textual data, where each sentence was split into individual words (tokens). The process started by cleaning the text, converting it to lowercase, and removing punctuation, symbols, and any unrecognized characters. This step is crucial as it ensures consistency and reduces noise in the data. Following the cleaning, the Tokenizer class from the Keras library was employed to convert the cleaned text into a sequence of integers, where each unique word from the vocabulary was mapped to a specific index. In this experiment, a vocabulary size of 2000 unique words was specified, meaning the tokenizer considered only the most frequent 2000 words in the dataset and ignored the less frequent words. This process simplifies the model while enhancing its ability to concentrate on the most important and relevant words. After tokenizing the text, stop words such as "and" or "is" that add little meaning were removed. A rule-based stemmer was then used to reduce words to their root form, allowing different versions of the same word, like "running" and "run," to be recognized as the same token.

Padding: Padding is a method used to make sure all input sequences are of equal length. This is especially important for neural networks, which need inputs to have consistent dimensions. In this experiment, after the text was tokenized into sequences of integers, the resulting sequences varied in length depending on the number of words in each input text. To ensure all inputs had the same shape, shorter sequences were padded to match the length of the longest one, while longer sequences were shortened when needed. This helped maintain uniformity for processing in neural networks. The maximum sequence length was set to 100 words in this experiment. Any sequence shorter than this length was padded with zeros at the beginning (pre-padding) to ensure uniformity across all input sequences. Padding guarantees that all sequences entering the model are of equal length, enabling efficient batch processing and preventing shorter sequences from negatively impacting the learning process. By padding the sequences to a fixed length, the model could focus on important patterns in the data without being affected by variations in the lengths of different texts. This method also helps avoid potential bias caused by varying input sequence lengths during the model's training and testing phases.

Word Embedding: Word embedding plays an important role in processing Natural Language Processing (NLP) by representing words or phrases as dense vectors of real numbers. This technique captures the meaning of words by considering the context in which they appear within the text. In this experiment, word embedding was performed using the Word2Vec model from the gensim library. Word2Vec is a widely used technique for creating word embeddings, representing words within a continuous vector space. This approach allows words with similar meanings to have closely aligned vector representations. Unlike traditional one-hot encoding, which uses sparse binary vectors, Word2Vec places words in a lower-dimensional space, capturing their semantic relationships by analyzing how often they appear together in the text. For this experiment, a 100-dimensional Word2Vec model was trained using the cleaned text data to produce word embeddings. Every word in the vocabulary was represented by a 100-dimensional vector, with each dimension reflecting various aspects of the word's meaning. The Word2Vec model uses a sliding window approach, where it predicts a word based on its surrounding context (window size), learning meaningful representations. This helped capture the relationships between words that frequently appear together, such as synonyms or related concepts. Once the Word2Vec model was trained, an embedding matrix was created, where each word in the vocabulary was mapped to its corresponding vector from the Word2Vec model.

*2.3 Proposed Hybrid Model*

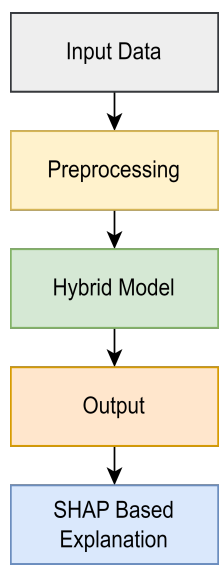

**Fig. 1.** Pipeline of the proposed system

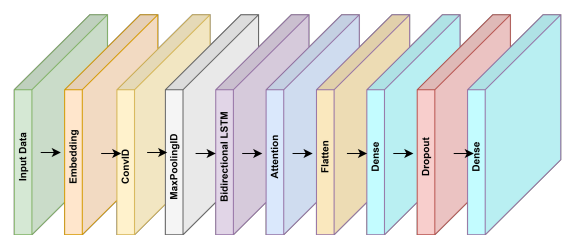

**Fig. 2.** CNN-BiLSTM model architecture

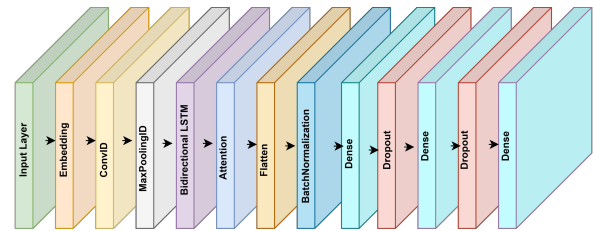

**Fig. 3.** Fine-tuned CNN-BiLSTM architecture

Figure 1 represents the pipeline of the proposed system. Initial model used to detect suicidal ideation was a CNN-BiLSTM hybrid model, leveraging the advantages of Convolutional Neural Networks (CNN) along with Bidirectional Long Short-Term Memory (BiLSTM) layers, illustrated in figure 2. This combination aimed to enhance performance by capturing both local patterns and long-range dependencies in the text. This hybrid architecture was specifically chosen to leverage the ability of CNNs to extract local patterns from text data and the capacity of BiLSTMs to capture long-term dependencies within sequences, which is crucial for understanding the context in social media posts. The following provides a detailed breakdown of each layer in the initial model and explains how each part plays a role in the overall operation of the neural network.

The model starts with a word embedding layer, which converts raw text into dense vector representations that reflect the meanings of words. This layer uses the Word2Vec technique, pre-trained on a large text corpus. Word embeddings enable the model to recognize relationships between words by analyzing their context, which is essential for natural language processing tasks. By representing words in a continuous vector space, the embedding layer provides more compact and meaningful input compared to older methods like one-hot encoding.

Next, the data is passed through a Convolutional Neural Network (CNN) layer, which is essential for capturing local dependencies within the text data. In this layer, the model applies multiple convolutional filters to the input sequences, where each filter detects a specific feature, such as common n-grams or word patterns. The filters slide across the sequence, generating feature maps that highlight the presence of these local patterns. When detecting suicidal ideation, this may involve identifying patterns such as phrases or expressions that are often linked to suicidal thoughts. The CNN layer in this model uses 128 filters with a kernel size of 5, which allows it to capture meaningful word groups while maintaining computational efficiency. Following the convolution step, a max-pooling layer is used to decrease the dimensionality of the feature maps by extracting the most significant features. This pooling process allows the model to concentrate on the most essential parts of the data, while also minimizing the chance of overfitting.

After the CNN layer, the model integrates a Bidirectional Long Short-Term Memory (BiLSTM) layer to capture long-term patterns within the text. Unlike traditional recurrent neural networks (RNNs) that often encounter issues like the vanishing gradient problem, LSTMs utilize components such as forget and input gates. These mechanisms enable the model to preserve important information across longer sequences, ensuring that relevant context is maintained. In this case, the bidirectional aspect means that the model processes the input sequence from both directions—forward and backward—thereby allowing it to consider the entire context of the text, including both preceding and succeeding words in a sentence. This is particularly useful for suicidal ideation detection because the meaning of a sentence can drastically change based on the surrounding context. For example, the presence of negations or conditional phrases (e.g., "I feel like I don't want to live anymore, but…") can only be understood when the entire sequence is taken into account. The BiLSTM layer in this model consists of 64 units, balancing complexity and computational efficiency.

To further refine the output of the BiLSTM, the model introduces an attention mechanism, which is applied after the BiLSTM layer. Attention mechanisms enable the model to concentrate on the most critical parts of the sequence, improving its ability to make accurate predictions. Instead of giving equal importance to every part of a sequence, attention assigns varying weights to different words or phrases, emphasizing those that are most relevant to the specific task. In this model, attention allows the network to concentrate on certain parts of a social media post that may better reflect signs of suicidal ideation, such as emotionally charged words or direct references to suicidal thoughts. This layer is crucial for improving the interpretability and accuracy of the model, as it ensures that critical information is not lost or diluted by the rest of the sequence.

After the attention mechanism identifies the key features, the output is sent to a flattening layer, which transforms the two-dimensional data from the earlier layers into a one-dimensional vector. This step readies the data for processing by the fully connected layers. Once flattened, the vector moves through a dense layer containing 64 neurons and activated by the Rectified Linear Unit (ReLU) function. This layer plays a key role in learning non-linear combinations of the high-level features identified by the CNN and BiLSTM layers. The ReLU activation adds non-linearity to the model, allowing it to detect complex patterns within the data an important factor in differentiating between suicidal and non-suicidal posts.

To reduce the risk of overfitting, a dropout layer is placed after the dense layer, with a dropout rate of 0.5. This layer functions by randomly deactivating a portion of the input units during training. By doing so, the model is encouraged to develop stronger and more generalized features, preventing it from depending too heavily on any specific input. This regularization technique is particularly effective when working with complex models and large datasets, such as social media text data, where overfitting is a common risk.

The model concludes by generating predictions through a sigmoid activation layer, which outputs a probability score ranging from 0 to 1. This score reflects the likelihood that a social media post contains signs of suicidal ideation. As this is a binary classification task (suicidal vs. non-suicidal), the sigmoid activation is well-suited for the job, as it transforms the network's raw output into a probability that can be easily interpreted and used to make a final decision.

*2.4 Fine Tuning*

In the fine-tuned version of the CNN-BiLSTM model, shown in figure 3, several key improvements were introduced, optimizing various components to enhance the model's overall performance. The core structure comprising convolutional layers, bidirectional LSTM units, and attention mechanisms remained the same, but crucial refinements were made to reduce overfitting, improve generalization, and achieve more stable training.

One of the most significant improvements in the fine-tuned model was the use of optimized regularization techniques, specifically the L2 regularization and dropout layers. The dropout rate in the fine-tuned version was set at 0.5 for the dense layers, similar to the original model, but now included L2 regularization (with a factor of 0.01) for the convolutional, BiLSTM, and dense layers. This type of regularization discourages large weights, allowing the model to generalize more effectively and reducing the risk of overfitting. The combination of dropout and L2 regularization ensured that the model could maintain its ability to learn from diverse examples without memorizing the training data, which significantly improved the model's robustness on unseen test data.

The introduction of Batch Normalization after the attention mechanism was another key refinement in the fine- tuned model. Batch normalization helps in normalizing the output of the previous layer, reducing internal covariate shifts and allowing the model to train faster and more stably. By introducing this step before the fully connected layers, the model's convergence was improved, leading to better overall accuracy.

Another major improvement was the introduction of early stopping during the training process. In the original model, training continued for a fixed number of epochs, even if the model had already reached its optimal performance. In the refined model, early stopping was used to track validation loss. Training automatically stopped if the validation loss did not improve for five consecutive epochs, and the model checkpoint saved the best-performing weights during this process. This adjustment not only helped prevent overfitting but also reduced training time, making the process more efficient and avoiding excessive training beyond the optimal point.

The fine-tuned model also benefited from learning rate optimization. The learning rate was set to a low value (0.0001) in the fine-tuned model, allowing smaller and more precise updates to the model's weights. This lower learning rate prevented the model from overshooting the optimal weight values during training, which is particularly important in later epochs when fine-tuning is crucial for high accuracy. The Adam optimizer was used to manage these updates efficiently, ensuring smooth convergence and minimal fluctuations in the validation performance.

*2.5 Performance Measure*

Accuracy Score: It measures the ratio of correctly classified comments out of the total number of comments in the test dataset, represented in eq. (1). It's a simple and commonly used metric to evaluate classification performance.

$$\text{Accuracy} = \frac{\text{TP} + \text{TN}}{\text{TP} + \text{PP} + \text{TN} + \text{FN}} \quad (1)$$

True Positives (TP), True Negatives (TN), False Positives (FP), False Negatives (FN): These components collectively constitute the confusion matrix, offering a more comprehensive perspective on the classification performance. TP and TN represent correct classifications, while FP and FN indicate misclassification.

Precision: It measures the model's capability to accurately recognize a specific class. Precision can be formulated or defined as showed in eq. (2):

$$\text{Precision} = \frac{\text{TP}}{\text{TP} + FP} \quad (2)$$

Recall (Sensitivity): It calculates the ratio of actual positives that were accurately predicted by the model. Thus, recall can be formulated as eq. (3):

$$\text{Recall} = \frac{\text{TP}}{\text{TP} + FN} \quad (3)$$

F1-Score: It is the harmonic mean of precision and recall. It considers both precision and recall, resulting in a lower value compared to accuracy. The formula for calculating the F1-score is represented in eq. (4):

$$\text{F1} = \frac{2 \times \text{Precision} \times \text{Recall}}{\text{Precision} + \text{Recall}} \quad (4)$$

Confusion Matrix Visualizations: Graphical representations illustrating the performance of classifiers in terms of correct and incorrect classifications.

## 3. Results and discussion
*3.1 Results*

To evaluate the proposed CNN-BiLSTM hybrid model, the dataset was divided into training, validation, and testing sets in an 80:10:10 ratio. The model was trained over 40 epochs with a batch size of 512, utilizing the Adam optimizer. Early stopping was applied with a patience of 4 epochs to avoid overfitting, and the learning rate was set at 0.0001.

The CNN-BiLSTM model performed exceptionally well throughout both training and evaluation. During training, it achieved a training accuracy of 98.54%, demonstrating its ability to correctly classify nearly all examples from the training set. Notably, the model also recorded a validation accuracy of 94.46%, reflecting its capacity to generalize effectively to new, unseen data. The close match between training and validation accuracy suggests the model avoided overfitting and successfully identified the key patterns within the data.

Once training was complete, the model was evaluated on a completely unseen test set. It achieved a test accuracy of 94.29%, confirming that the model maintained its strong performance when applied to new data. However, accuracy alone does not fully reflect the model's effectiveness in detecting suicidal ideation, so additional performance metrics were calculated for a more comprehensive evaluation.

The precision of the model was 0.9458, indicating that 94.58% of the model's positive predictions (instances classified as indicating suicidal ideation) were correct. Precision is a particularly important metric for this task, as misclassifying non-suicidal instances as suicidal (false positives) could lead to unnecessary concerns or interventions.

The model achieved a recall of 0.9400, indicating it correctly identified 94.00% of the true cases of suicidal ideation in the dataset. This high recall is essential, as failing to detect actual cases (false negatives) could lead to serious consequences. Additionally, the model recorded an F1 score of 0.9429, which reflects a strong balance between precision and recall. This score highlights the model's effectiveness in accurately detecting true cases while keeping false positives and false negatives to a minimum.

The Confusion Matrix and ROC Curve are valuable tools for evaluating a model's performance. The Confusion Matrix shows (represented in figure 4) the counts of true positives, true negatives, false positives, and false negatives, providing insight into how well the model classifies each category. This information is used to calculate important metrics such as precision, recall, and F1-score, which reflect the model's ability to detect positive cases (sensitivity) and negative cases (specificity). The ROC Curve (Receiver Operating Characteristic Curve) plots the true positive rate against the false positive rate, illustrating the balance between sensitivity and specificity at various thresholds. The model's effectiveness is measured by the Area Under the Curve (AUC), where a score near 1 indicates strong performance, while an AUC around 0.5 suggests the model performs no better than random guessing. By using both the Confusion Matrix and ROC Curve, it becomes easier to assess the model's accuracy, dependability, and overall strength in making predictions.

The model achieved a strong AUC (Area Under the Curve) score of 0.9851, showed in figure 5. A high AUC score indicates the model's excellent ability to distinguish between positive (suicidal ideation) and negative (non-suicidal) classes across various decision thresholds. This high AUC value demonstrates that the model is reliable at ranking true positive samples higher than negative samples, making it well-suited for applications where distinguishing between the two classes is critical.

These results indicate that the fine-tuned CNN-BiLSTM model is highly effective at identifying suicidal ideation in text, achieving strong performance metrics and generalizing well to unseen data.

These results demonstrate a significant improvement over the initial model performance, which had an accuracy of 92.81%. The fine-tuning and early stopping techniques were effective in optimizing the model's parameters and enhancing its generalization capability.

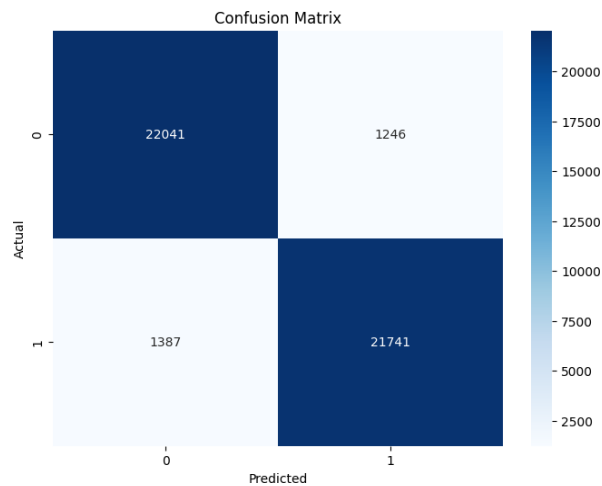

**Fig. 4.** Confusion matrix for the proposed model's prediction

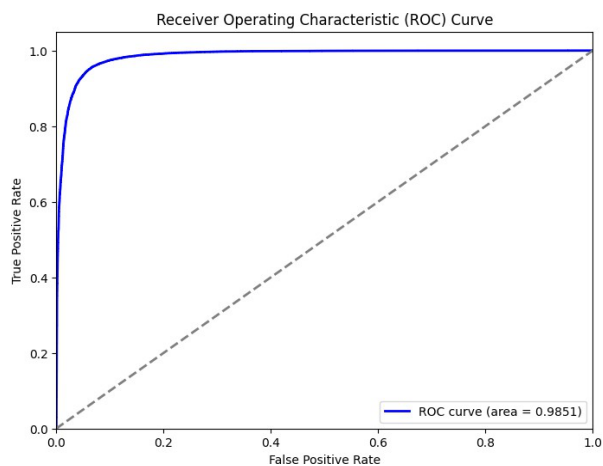

**Fig. 5.** ROC curve for the proposed model

*3.2 Discussion*

The proposed CNN-BiLSTM hybrid model with an attention mechanism effectively captures both local features and long-term dependencies in the text data. The convolutional layers extract important n-gram features, while the BiLSTM layers capture the sequential context of the text. The attention mechanism further enhances the model by allowing it to focus on the most relevant parts of the input text.

The evaluation metrics demonstrated the success of these improvements. The fine-tuned model achieved an ac- curacy of 94.29%, a significant improvement over the original 92.81%. Moreover, key metrics such as recall and F1-score also showed marked improvements, reflecting the model's enhanced ability to correctly identify suicidal ideation while minimizing false positives and negatives. These improvements were particularly crucial given the sensitive nature of the task, where accurately detecting suicidal tendencies is of utmost importance.

*3.2.1 Explainability with SHAP*

A significant aspect of this study is the use of explainability techniques to interpret the model's predictions. In this case, SHAP (SHapley Additive exPlanations) was applied to provide insight into how the model arrived at its decisions. SHAP is a valuable method that explains the predictions of machine learning models by assigning an importance score to each feature based on its contribution to a specific outcome. Rooted in game theory, SHAP offers a consistent way to measure feature importance by estimating how much each input affects the final prediction. In this study, SHAP helped identify which words or tokens played the largest role in determining whether a piece of text suggested suicidal ideation. This level of interpretability not only enhances trust in the model but also provides deeper insight into the factors driving its classifications.

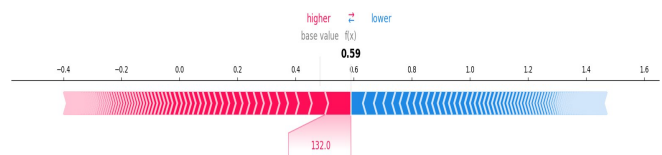

**Fig. 6.** SHAP force plot

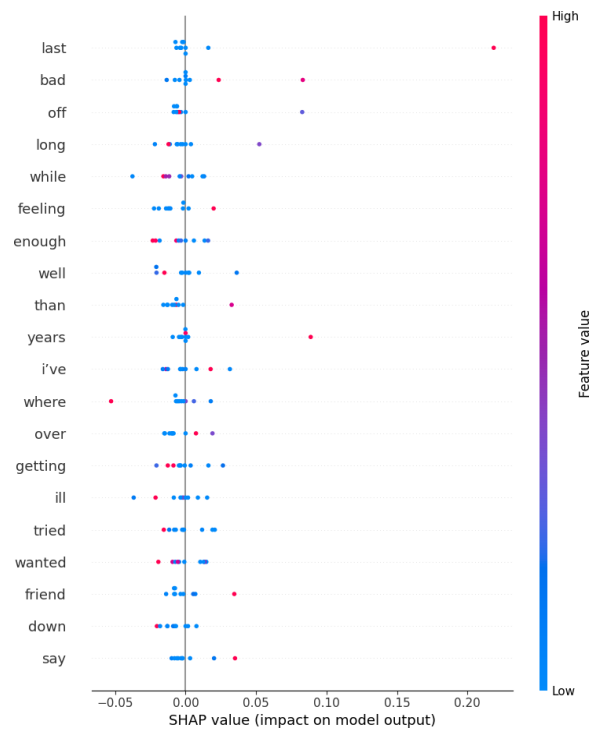

**Fig. 7.** SHAP summary plot

*3.2.2 Local explanation: Force plot*

Figure 6 illustrates a SHAP force plot, which provides a detailed, instance-specific explanation for a single prediction. The base value represents the average model output (i.e., the average prediction probability before considering specific features). As individual features (words) from the input text are processed, they either increase or decrease the prediction probability, moving the model's output away from or towards the base value.

Features marked in red represent words that increased the model's prediction, pushing it towards a higher likelihood of suicidal ideation. Conversely, features in blue decreased the prediction probability. In this specific instance, the model predicted a probability of 0.59 for suicidal ideation. The word contributing most significantly to this prediction is highlighted in red, suggesting that its presence increased the likelihood of suicidal ideation.

*3.2.3 Global explanation: Summary plot*

Figure 7 shows a SHAP summary plot, providing an overall view of how various words influenced the model's pre- dictions across the dataset. Each dot represents a specific word (feature) from the data, with the x-axis displaying the SHAP value, which reflects the word's impact on the model's decision. Words with higher SHAP values increase the likelihood of the model predicting suicidal ideation, whereas lower or negative SHAP values decrease this probability. This visualization helps illustrate the key factors driving the model's outputs.

In the summary plot, words like "last", "bad", and "long" are shown to have the highest positive SHAP values, meaning their presence strongly contributes to predictions of suicidal ideation. The color gradient of each dot represents the feature value, with red signifying higher feature values and blue indicating lower ones. This visual cue helps convey the relationship between the feature's magnitude and its influence on the model's predictions. For instance, a high SHAP value for the word "last" suggests that when it appears in a certain context (e.g., high feature value), it significantly increases the likelihood of a prediction for suicidal ideation.

*3.2.4 Interpretation and significance*

By employing SHAP, we gain both local and global insights into the model's behavior. The force plot provides a localized view, allowing us to understand how specific words in a single post influenced that particular prediction. This instance-specific interpretability is essential for real-world applications where it is important to justify individual predictions. The summary plot, on the other hand, offers a broader perspective by showing the overall contribution of each word to the model's decisions across all posts in the dataset.

The use of SHAP ensures that the model's predictions are not only accurate but also transparent and explainable. In the context of suicidal ideation detection, where every decision must be carefully validated, SHAP provides a critical layer of interpretability. By visualizing which words push the model towards or away from identifying suicidal ideation, SHAP allows for a greater degree of trust and reliability in the model's predictions.

SHAP-based explainability strengthens the model's applicability in the detection of suicidal ideation, providing a clearer understanding of how decisions are made. This interpretability is not only vital for model validation but also necessary for ethical considerations in deploying such models for mental health monitoring.

*3.2.5 Comparison with existing literature*

In this subsection, the performance of the proposed CNN-BiLSTM hybrid model with an attention mechanism is compared against various state-of-the-art models reported in the literature for suicidal ideation detection. The comparison focuses on the accuracy achieved by each model, providing a clear indication of the effectiveness of this approach relative to existing methods. The models selected for comparison include transformer-based models, ensemble learning techniques, and other deep learning architectures commonly used in this domain. The results are summarized in Table 1, highlighting the superior performance of our proposed model.

**Table 1**
Comparison of model performance for suicidal ideation detection

| Reference | Model Type | Accuracy (%) |
|---|---|---|
| **Proposed Model** | **CNN + BiLSTM + Attention** | **94.29** |
| Jain et al. [9] | RF | 77.30 |
| Tadesse et al. [10] | CNN + LSTM | 93.80 |
| Sawhney et al. [11] | Transformer | 82.27 |
| Naghavi et al. [12] | Ensembled DT | 93.00 |
| Schoene et al. [21] | Transformer + BiLSTM | 87.34 |

This table provides a clear comparison of the model's performance against other well-known models in the field, illustrating the effectiveness of the proposed approach in accurately detecting suicidal ideation from social media texts.

## 4. Conclusion

The study introduced a CNN-BiLSTM hybrid model with an attention mechanism for detecting suicidal ideation from social media texts, showcasing significant results. By combining CNNs for capturing local textual patterns and BiLSTM for understanding broader context, enhanced by attention for focusing on key text segments, the model achieved notable predictive accuracy. Optimization through fine-tuning and early stopping improved accuracy from 92.81% to 94.29%, highlighting reduced overfitting and better generalization. The application of SHAP as an Explainable AI technique added valuable interpretability, enabling insights into the model's decision-making process, thus promoting trust in its application for mental health monitoring. Despite SHAP's interpretability benefits, it has limitations in mental health contexts: SHAP values represent statistical patterns, not clinical insights, and practitioners might misinterpret feature importance as clinical significance. It is important to acknowledge that this study focuses on technical validation using publicly available social media datasets and does not include clinical or real-world validation with mental health professionals.

Future research should address challenges unique to suicidal ideation detection in clinical practice. Developing temporal models that track risk escalation patterns over time could identify users transitioning from ideation to active planning, while integrating contextual signals such as posting time patterns, social isolation indicators, and changes in linguistic markers could improve early detection of crisis states. Culturally-sensitive models are essential, as suicidal ideation expression varies significantly across cultures, requiring detection of culture-specific metaphors and indirect distress signals often missed by current models. Clinical validation with mental health professionals must assess whether SHAP explanations align with clinical reasoning and support intervention decisions. Additionally, ethical frameworks for handling false positives, managing user consent in crisis situations, and establishing appropriate intervention thresholds based on available mental health resources would be essential for responsible deployment. These domain-specific advances would move beyond technical improvements to address the practical and ethical complexities of automated mental health monitoring.

**Acknowledgement**

This work is supported by PGRS2303109 and UMPSA Research Grant 2023 (Grant No. RDU230353).